\documentclass[11pt]{article}

\usepackage{ACL2023}

\usepackage{times}
\usepackage{latexsym}
\usepackage{graphicx}
\usepackage[T1]{fontenc}

\usepackage[utf8]{inputenc}
\usepackage{microtype}
\usepackage{algorithm}
\usepackage{algorithmic}
\usepackage{colortbl}
\usepackage{inconsolata}
\usepackage{multirow}  
\usepackage{amssymb}   
\usepackage{graphicx}  
\usepackage{amsmath}
\usepackage{color}
\usepackage{subfigure} 
\usepackage{makecell} 
\usepackage{booktabs} 
\usepackage{enumitem} 
\usepackage{multirow} 
\usepackage{booktabs} 
\usepackage{stfloats}
\usepackage{cite}
\usepackage{inconsolata}

\title{SANTA: Separate Strategies for Inaccurate and Incomplete \\ Annotation Noise in Distantly-Supervised Named Entity Recognition}

\author{Shuzheng Si$^{1, 2}$\footnotemark[1] , Zefan Cai$^{1,2}$\footnotemark[1] , \textbf{Shuang Zeng}$^{1,2}$, \\ 
\textbf{Guoqiang Feng}$^{1,2}$, \textbf{Jiaxing Lin}$^{1,2}$ and \textbf{Baobao Chang}$^{1}$\footnotemark[2] 
\\
$^1$National Key Laboratory for Multimedia Information Processing, Peking University \\ 
$^2$School of Software and Microelectronics, Peking University, China \\
\texttt{\{sishuzheng\}@stu.pku.edu.cn}
}

\begin{document}
\maketitle

\begin{abstract}
Distantly-Supervised Named Entity Recognition effectively alleviates the burden of time-consuming and expensive annotation in the supervised setting.
But the context-free matching process and the limited coverage of knowledge bases introduce inaccurate and incomplete annotation noise respectively.
Previous studies either considered only incomplete annotation noise or indiscriminately handle two types of noise with the same strategy. 
In this paper, we argue that the different causes of two types of noise bring up the requirement of different strategies in model architecture.
Therefore, we propose the \textbf{SANTA} to handle these two types of noise separately with 
(1) Memory-smoothed Focal Loss and Entity-aware KNN to relieve the entity ambiguity problem caused by inaccurate annotation, and (2) Boundary Mixup to alleviate decision boundary shifting problem caused by incomplete annotation and a noise-tolerant loss to improve the robustness.
Benefiting from our separate tailored strategies, we confirm in the experiment that the two types of noise are well mitigated.
SANTA also achieves a new state-of-the-art on five public datasets.



\end{abstract}
\renewcommand{\thefootnote}{\fnsymbol{footnote}}
\footnotetext[1]{~Equal Contribution.}
\footnotetext[2]{~Corresponding Author.}
\renewcommand{\thefootnote}{\arabic{footnote}}

\section{Introduction}
\label{intro}
\begin{figure}
    \centering
    \includegraphics[width=7.5cm]{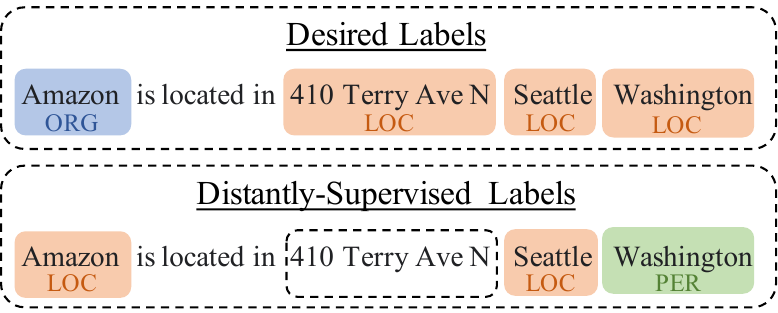}
    \caption{A sample generated by Distant Supervision. ``Amazon'' and ``Washington'' are inaccurate annotations. 
    ``410 Terry Ave N" is the incomplete annotation.}
    \label{fig_example}
\end{figure}

\noindent
As a fundamental task in NLP, Named Entity Recognition (NER) aims to locate and classify named entities in text, which plays an important role in many tasks such as knowledge graph construction \citep{peng-etal-2022-smile, li-etal-2022-self} and relation extraction \citep{zeng-etal-2021-sire, wang-etal-2022-learning-robust}.
To alleviate the burden of annotation in the supervised setting, Distantly-Supervised Named Entity Recognition (DS-NER) is widely used in real-world scenarios. 
It can automatically generate labeled training data by matching entities in existing knowledge bases with snippets in plain text.
However, DS-NER suffers from two inherent issues which introduce many noisy samples: (1) \textbf{inaccurate annotation}: the entity with multiple types in the knowledge bases may be labeled as an inaccurate type in the text, due to the context-free matching process, and (2) \textbf{incomplete annotation}: the knowledge bases with limited coverage of entities cannot label all entities in the text.
As shown in Figure \ref{fig_example}, the entity types of ``Amazon" and ``Washington" are wrongly labeled owing to context-free matching, and ``410 Terry Ave N" is not recognized due to the limited coverage of knowledge bases.

Due to the sensitivity to the noise, the original supervised methods achieve poor performances in DS-NER.
Therefore, many works have been proposed to handle the issue.
Some works attempted to focus on solving incomplete annotation noise in DS-NER, including positive-unlabeled (PU) learning \citep{DBLP:conf/acl/PengXZFH19,zhou-etal-2022-distantly}, negative sampling \citep{DBLP:conf/iclr/LiL021,li-etal-2022-rethinking}, and retrieval augmented inference with contrastive learning \citep{si-etal-2022-scl}.
However, the ignorance of inaccurate annotation noise limits the model to further improve the performance.
Recently, \citet{zhang-etal-2021-improving-distantly} jointly trained two teacher-student networks to handle the two types of noise with the same strategy.
\citet{meng-etal-2021-distantly} adopted a noise-robust learning scheme and self-training for the whole training set to avoid overfitting in noise.
However, by handling both types of noise with the same strategy, these methods failed to address the unique characteristics of different type of noise, thereby limiting their ability to effectively handle the noise in DS-NER.

As the causes of the two types of noise are different, both of two noise may lead to different problems in DS-NER task.
Inaccurate annotation noise in training data can lead to serious entity ambiguity problem in the DS-NER.
As exampled in Figure \ref{fig_example}, when a model is trained on data where ``Washington" is consistently labeled as ``PER" (person) due to context-free matching, the model may continue to predict ``Washington" as a "PER" even in contexts where it should be labeled as a ``LOC" (location), such as when it refers to the city of Washington.
Incomplete annotation noise can lead to the decision boundary shifting problem \citep{si-etal-2022-scl} in the DS-NER task.
This problem occurs when the model is trained on data where some entity spans are not labeled, causing the model to shift its decision boundary, making it more likely to predict an entity span as a non-entity type. 
Therefore, the noise in the spans labeled as entities by distant supervision leads to the ambiguity problem, and the noise in the spans labeled as non-entities leads decision boundary shifting problem.
To further improve the performance in DS-NER, we argue that the two types of noise should be handled separately with specialized designs in model architecture.
This can help the model to address the specific problems posed by each type of noise and lead to the better overall performance of the model.

 
In this paper, we propose the \textbf{S}eparate str\textbf{A}tegies for i\textbf{N}accurate and incomple\textbf{T}e \textbf{A}nnotation noise in DS-NER (\textbf{SANTA}).
Unlike previous works in DS-NER, we introduce different strategies to handle the two types of noise respectively.
For inaccurate annotation, we propose Memory-smoothed Focal Loss (MFL) and Entity-aware KNN.
These strategies aim to address the entity ambiguity problem posed by inaccurate annotation noise.
For incomplete annotation, we propose Boundary Mixup to handle decision boundary shifting problem caused by incomplete annotation, which generates augmented instances by combining the instances around the boundary and the entity instances.
Due to further training on the augmented instances, the biased decision boundary can be pushed towards right (fully supervised) side as the augmented instances exist between the biased boundary and the right boundary.
Meanwhile, we empirically analyze the characteristics of incomplete annotation noise, then adopt a noise-tolerant loss to further improve the model's robustness to this type of noise.

Experiments show SANTA achieves state-of-the-art on five public DS-NER datasets.
Further analysis shows the effectiveness of each designed module and separate handling.

\section{Related Work}
\label{related_work}
\noindent
To address the data scarcity problem, several studies attempted to annotate datasets via distant supervision.
Using external knowledge bases can easily get training data through string matching, but introduces two issues: inaccurate annotation noise and incomplete annotation noise.
To address these issues, various methods have been proposed.

\paragraph{Only Focusing on Incomplete Annotation.}
Several studies \citep{shang-etal-2018-learning, DBLP:conf/coling/YangCLHZ18, jie-etal-2019-better} modified the standard CRF to get better performance under the noise, e.g., Partial CRF.
LRNT \citep{DBLP:conf/emnlp/CaoHCLJ19} leaved training data unexplored fully to reduce the negative effect of noisy labels.
AdaPU \citep{DBLP:conf/acl/PengXZFH19} employed PU learning to obtain unbiased estimation of the loss value. 
Furthermore, Conf-MPU \citep{zhou-etal-2022-distantly} used multi-class PU learning to further improve the performance.
\citet{DBLP:conf/iclr/LiL021} performed uniform negative sampling to mitigate the misguidance from unlabeled entities. 
\citet{li-etal-2022-rethinking} then proposed a weighted sampling distribution to introduce direction to incomplete annotation when negative sampling.
\citet{si-etal-2022-scl} adopt supervised contrastive-learning loss and retrieval-augmented inference to mitigate the decision boundary shifting problem.
However, these studies only addressed incomplete annotation noise, ignoring inaccurate annotation noise, which also exists in DS-NER. 

\paragraph{Handling Two Types of Noise With the Same Strategy.}
To further explore the information in DS-NER text, many studies attempted to consider both inaccurate and incomplete annotation noise.
BOND \citep{DBLP:conf/kdd/LiangYJEWZZ20} designed a teacher-student network to drop unreliable labels and use pseudo labels to get more robust model.
SCDL \citep{zhang-etal-2021-improving} further improved the performance by jointly training two teacher-student network and refining the distant labels.
RoSTER \citep{meng-etal-2021-distantly} adopt a noise-robust learning scheme and self-training to improve the robustness.
CReDEL \citep{ying-etal-2022-label} trained an automatic distant label refinement model via contrastive learning as a plug-in module for other DS-NER models.
Although these works jointly considered the two types of noise, they did not take into account the difference between the two types of noise.
Due to the different causes and frequencies of the two different types of noise, we argue that they should be handled separately with specialized strategies, which have not been considered in previous work.

\begin{figure*}
    \centering
    \includegraphics[scale=0.5]{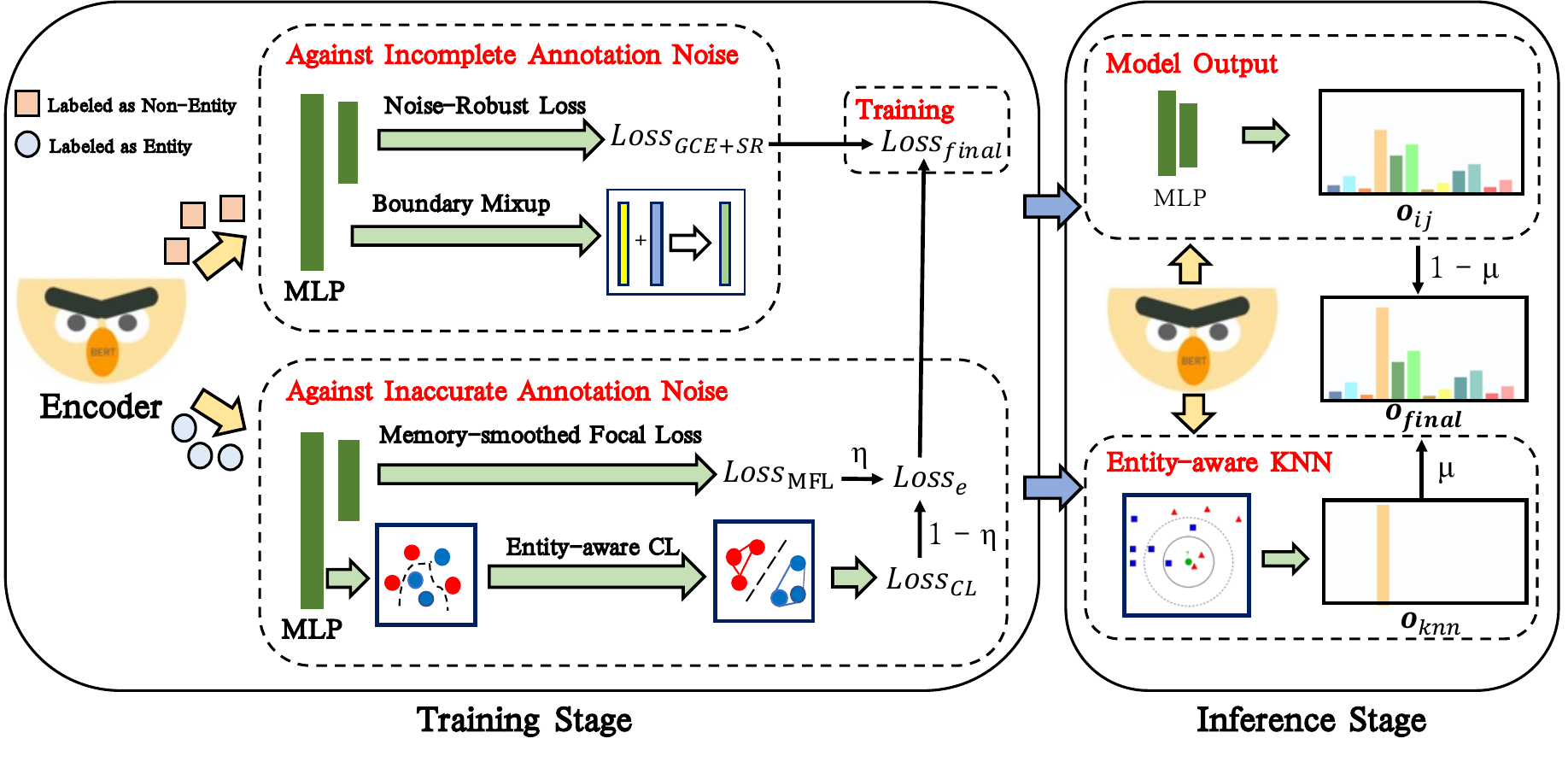}
    \caption{General architecture of SANTA.}
    \label{fig_model}
\end{figure*}

\section{Method}
As shown in Figure \ref{fig_model}, SANTA separately handles the spans labeled as entities and labeled as non-entities.
SANTA uses the MFL and Entity-aware KNN to address the entity ambiguity problem caused by inaccurate annotation.
SANTA adopts noise-tolerant GCE + SR loss and Boundary Mixup to handle the incomplete annotation.

\subsection{Span-based NER Model}

We follow the same span-based NER model as \citet{li2022rethinking, li2020empirical} and \citet{si-etal-2022-scl}.
For sentence $[x_1,x_2,...,x_n]$, we use a pre-trained language model as an encoder to get the representations for every token $x$ in the sentence:
\begin{align}
[\textbf{h}_1,\textbf{h}_2,...,\textbf{h}_n] = {\rm Encoder}([\textbf{x}_1,\textbf{x}_2,...,\textbf{x}_n])
\end{align}
where $h_i$ is the representation for token $x_i$. 

For each span $s_{i,j}$ ranging from $i$-th token to $j$-th token, the span representation $\textbf{s}_{i,j}$ is calculated as:
\begin{align}
\textbf{s}_{i,j} = \textbf{h}_i \oplus \textbf{h}_j \oplus (\textbf{h}_i - \textbf{h}_j) \oplus (\textbf{h}_i \odot \textbf{h}_j)
\end{align}
where $\oplus$ is the concatenation operation and $\odot$ is the element-wise product operation. 

Then, we use linear layer and activation function to get a more dense representation:
\begin{align}
\label{W_dim}
\textbf{r}_{i,j} &= {\rm tanh}(W\textbf{s}_{i,j})
\end{align}

Finally, we can obtain the entity label distribution $o_{i,j}$ for every span $s_{i,j}$ as:
\begin{align}
\textbf{o}_{i,j} &= {\rm softmax}(V\textbf{r}_{i,j})
\end{align}
where $W$ and $V$ are trainable parameter.

\subsection{Against Inaccurate Annotation Noise}
We observe that the inaccurate annotation in spans labeled as entities leads to two problems:
(1) it can cause fluctuations in the training process and make it difficult to achieve consistent predictions;
(2) it can lead to the entity ambiguity problem.

\subsubsection{Memory-smoothed Focal Loss}


\noindent
In the training process, the inaccurately labeled entities and the similar true entities supervise the model back and forth and cause fluctuations in the model's learning.
For example, the ``Washington" in Figure \ref{fig_example} is inaccurately labeled as ``PER" (person), the model trained with it tends to predict ``Washington" as ``PER" instead of ``LOC" (location).
However, if the model is also exposed to similar and correctly labeled entities, such as ``Seattle" labeled as ``LOC", the model may also learn to generalize "Washington" as a "LOC". 
This back-and-forth supervision can make the model's performance being less consistent and less accurate. 
Intuitively, if we can smooth out such fluctuations, the performance of model can be further improved.

This observation motivates us to propose Memory Label Smoothing (MLS) to reduce the fluctuation.
MLS memorizes model predictions to continuously update the soft labels during the training stage.
The memorized soft labels are used to smooth the training of every sample, which tries to enable the model jointly consider the label information from previous predictions, so that learning for a particular sample does not fluctuate tremendously.
Given each span $x_{i,j}$ labeled as entity, if its prediction is correct, the soft label corresponding to the entity type $y_{i,j}$ will be updated using the model output $o_{i,j}$.
Specifically, in training total $T$ epochs, we construct $\hat{Y} = [\hat{Y}^0, ..., \hat{Y}^t, ..., \hat{Y}^T ]$ as the memory soft labels at different training epochs.
$\hat{Y}^t$ is a matrix with $|L|$ rows and $|L|$ columns, 
and each column in $\hat{Y}^t$ corresponds to the vector with latitude $|L|$ as soft label for one category. 
$L$ denotes the set of entity types.
For epoch $t>0$, $\hat{Y}^t$ is defined as:
\begin{align}
\hat{Y}_{y_{i,j},l}^{t} = \frac{1}{N}\sum_{l \in L} \mathbb I\{l=y_{i,j}\}o_{i,j}
\end{align}
where $N$ denotes the number of correctly predicted entities
with label $y_{i,j}$.
$\mathbb I$ is indicator function.
$\hat{Y}^0$ is initialize as identity matrix.

Then the updated soft labels will be utilized to supervise the model in the next epoch and smooth the learning curve of the model.
In training epoch $t$, given a span $(x_{i,j},y_{i,j})$, we use the previous $G$ soft label $[\hat{Y}^{t-G},...,\hat{Y}^{t-1}]$ to supervise the model:
\begin{align}
Y_{final}^{t} = \sum_{l \in L} (\lambda \mathbb I\{l=y_{i,j}\} + (1-\lambda)\frac{1}{G}\sum_{g=1}^{G}\hat{Y}_{y_{i,j},l}^{t-g})
\label{equation:loss_mfl_pre}
\end{align}
where $G$ and $\lambda$ are hyperparameters. For $t<G$, we use all the previous soft labels to get $Y_{final}^{t}$.

Meanwhile, we adopt Focal Loss \citep{DBLP:journals/pami/LinGGHD20} to handle the entity ambiguity problem, which can be calculated as:
\begin{align}
\mathcal L_{FL} = - \sum_{l \in L} \alpha(1 - {o}_{i,j})^{\gamma} \log ({o}_{i,j})
\end{align}
where  $\alpha$ and $\gamma$ are hyperparameters. It can be seen that Focal Loss has a greater weight for ambiguous samples with low confidence.

Combined with the capability from MLS to reduce the fluctuation and the capability from Focal Loss to handle the ambiguous samples, the Memory-smoothed Focal Loss (MFL) can be defined as:
\begin{small} 
\begin{align}
\scriptsize
\mathcal L_{MFL} = \sum_{x_{i,j} \in D_e}\sum_{l \in L} \alpha{(Y_{final}^{t})}_l(1 - {o}_{i,j})^{\gamma} \log ({o}_{i,j})
\label{equation:loss_mfl}
\end{align}
\end{small}
\quad In addition, MFL is only performed on spans labeled as entities $D_e$ to focus on the problems caused by inaccurate annotation.

\subsubsection{Entity-aware KNN}
The proposed MFL method may not completely solve the entity ambiguity caused by inaccurate annotation noise, and therefore relying solely on the output of the trained model may still result in limited performance. 
To address this issue, we propose Entity-aware KNN during the inference stage to facilitate the decoding process by retrieving similar labeled samples in the training set, which further improves the overall performance.

Entity-aware KNN consists of two parts, including Entity-aware Contrastive Learning (Entity-aware CL) and KNN-augmented Inference.
Specifically, we use Entity-aware CL to close the distance between the span labeled as entity with the same type, pull the distance of different types.
Therefore, we could get the better representation to easily use KNN-augmented Inference to get a retrieved distribution $o_{knn}$.
Finally, we interpolate the output $o_{i,j}$ from model with $o_{knn}$ to further handle the entity ambiguity problem.

To improve the performance of KNN retrieving, Entitiy-aware CL pulls spans belonging to the same entity type together in representation space, while simultaneously pushing apart clusters of spans from different entity types.
Therefore, the type of entities could be better distinguished.
We use the cosine similarity as metric between the representations 
$r_{i,j}$ and $r_{\hat i, \hat j}$ of span $x_{i,j}$ and $x_{\hat i, \hat j}$:

\begin{equation}
\label{cos}
    d_{s_{i,j},s_{\hat i, \hat j}} = \frac{\textbf{r}_{i,j} \cdot \textbf{r}_{\hat i,\hat j}}{\lvert \textbf{r}_{i,j} \rvert \lvert \textbf{r}_{\hat i,\hat j} \rvert}
\end{equation}

Then the Entity-aware CL $\mathcal L_{CL}$ is defined as:
\begin{gather}
\begin{small}
\label{cl}
\mathcal L_{CL} = -\sum_{l \in L_e}\sum_{r_{i,j} \in E_{l}} \frac{1}{N_l-1} \sum_{r_{\hat i,\hat j} \in E_{\bar{l}}} F(\textbf{r}_{i,j},\textbf{r}_{\hat i,\hat j})
\end{small}
\end{gather}
where $L_e$ is the entity label set; 
$N_{l}$ is the total number of spans with the same entity label $l$ in the batch; 
$E_{l}$ is the collection of all training spans with $l$-th entity label.  $F(\textbf{r}_{i,j},\textbf{r}_{\hat i,\hat j})$ is calculated as:
\begin{align}
F(\textbf{r}_{i,j},\textbf{r}_{\hat i,\hat j}) = \log \frac{\exp(d_{r_{i,j},r_{\hat i,\hat j}} / \tau) }{\sum_{r_{m,n} \in E_{\bar{l}}} \exp(d_{r_{i,j},r_{m,n}} / \tau)}
\label{equation:loss_cl}
\end{align}
where $\tau$ is the temperature. $E_{\bar{l}}$ is the collection of labeled entity spans not with entity label $l$.

After we get the easily distinguishable entity span representations, we propose KNN-augmented Inference to further relief the ambiguity problem by augmenting the trained model output.
KNN-augmented Inference can be split into three parts:

(i) Firstly, we cache each entity representation $r_{key}$ in training set and its label $l_{value}$ to construct a pair (key,value) $\in$ DataStore.

(ii) We calculate the cosine similarity as Eq. \ref{cos} between the representation $r_{i,j}$ from span $x_{i,j}$ and each cached representation from DataStore.

(iii) Then, we select the top $K$ most similar retrieved entities $D_K$ and then convert them into a one-hot distribution based on the KNN majority voting mechanism.
\begin{equation}
\begin{split}
y_{knn} = \arg\max_{l} &\sum_{D_{K}} \mathbb I(l_{value}=l), \forall l \in L \\
o_{knn} &= {\rm onehot}(y_{knn})
\end{split}
\end{equation}

(iv) Finally, we interpolate the ${o}_{i,j}$ from model with ${o}_{knn}$ to get the final distribution ${o}_{final}$:
\begin{align}
{o}_{final} = (1-\mu) * {o}_{i,j} + \mu * {o}_{knn}
\label{equation:loss_knn}
\end{align}
where $\mu$ is a hyperparameter to make a balance between two distributions.

In this way, we could use cached similar entities in the training set as memory to adjust the output of the trained model, therefore further mitigating the entity ambiguity problem.

\subsection{Against Incomplete Annotation Noise}
\noindent
We observe that the incomplete annotation in spans labeled as non-entities leads to two problems:
(1) the decision boundary shifting problem, 
and (2) the high asymmetric noise rate.

\begin{figure}
    \centering
    \includegraphics[width=7.5cm]{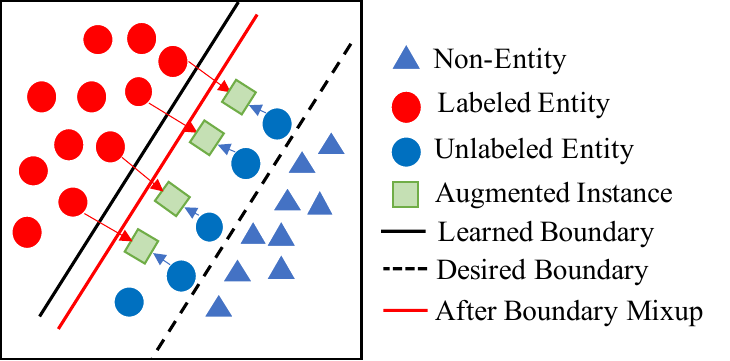}
    \caption{A toy case for decision boundary shifting problem and Boundary Mixup.}
    \label{fig_mixup}
\end{figure}

\subsubsection{Boundary Mixup}
\noindent
When the model is trained on data where some entity spans are not labeled, the learned decision boundary can be biased and the model tends to predict an entity span as a non-entity type.
As shown in Figure \ref{fig_mixup}, the learned decision boundary tends to shift from the fully supervised boundary (our desired boundary) towards the entity side.
Therefore, we propose Boundary Mixup to push the decision boundary to the unbiased (desired) side.

The model always makes a wrong prediction to the instances around the learned decision boundary due to the decision boundary shifting.
Motivated by this, if we can find the instances around the learned decision boundary, we can further find the location of the learned decision boundary.
Then we can utilize Mixip \citep{DBLP:conf/iclr/ZhangCDL18} to generate the augmented instances to modify the location of the learned decision boundary.
Specifically, if the span $x_{i,j}$ is predicted as non-entity $l_n$ with a confidence lower than $\varepsilon$, it may actually be an instance around the learned decision boundary.
Therefore, we randomly sample a entity $e$ with the most possible entity label $l_e$ according to $o_{i,j}$.
Then use Mixup between $(r_e,l_e)$ and $(r_{i,j},l_{i,j})$ to generate an augmented instance $(\hat r,\hat y)$:
\begin{align}
(\hat r, \hat y) &= (\theta' r_{i,j} + (1 - \theta')r_{e},\theta' l_{n} + (1 - \theta')l_{e})
\end{align}
where $\theta'$ is calculated as:
\begin{align}
\theta' &= {\rm max}(\theta , 1-\theta)\\
\theta \sim &{\rm Beta}(\alpha', \alpha'), \alpha' \in (0,\infty)
\label{equation:loss_mixup}
\end{align}
where $\alpha'$ is a hyperparameter.

As shown in Figure \ref{fig_mixup}, due to further training on the augmented instances, the biased decision boundary can be pushed towards the fully supervised side, mitigating the decision boundary shifting problem.

\subsubsection{Noise-Tolerant Loss}
To study the characteristics of incomplete annotation noise, we conduct an empirical analysis on BC5CDR and CoNLL2003.
We use the knowledge bases provided by \citet{zhou-etal-2022-improving} to relabel the two training sets using string matching, then compare the matching results with original well-labeled training sets to get the rate of inaccurate and incomplete annotation for different entity types.
As shown in Table \ref{tab_two}, incomplete annotation noise accounts for the majority of the noise in DS-NER as the Incomplete Rate is much higher than Inaccurate Rate.
Therefore, it is important to improve the robustness to incomplete annotation noise in DS-NER to improve the overall performance as the models are easy to overfit the noise.

Recently, \citet{meng-etal-2021-distantly} adopts generalized cross entropy (GCE) \citep{DBLP:conf/nips/ZhangS18} instead of cross entropy (CE) to improve the robustness in DS-NER. 
GCE calculated on spans labeled as non-entities $D_{n}$ can be described as following: 
\begin{align}
\mathcal{L}_{\mathrm{GCE}} =  \sum_{x_{i,j} \in D_{n}} \frac{1-o_{i,j}^{q}}{q}
\label{equation:loss_gce}
\end{align}
where $q$ denotes a hyperparameter;
when $q\rightarrow1$ GCE approximates Mean Absolute Error Loss (MAE), which is widely used in regression task; when $q\rightarrow0$, GCE approximates CE (using L’Hôpital’s rule).
\citet{DBLP:conf/aaai/GhoshKS17} theoretically proved that a loss function that satisfies the symmetric condition would be inherently tolerant to symmetric label noise.
However, the derived loss functions according to this design principle such as MAE, suffer from the underfitting \citep{DBLP:conf/icml/CharoenphakdeeL19}. 
GCE attempts to perform the trade-off between CE and symmetric loss MAE to improve the robustness.
Meanwhile, \citet{DBLP:conf/iccv/ZhouLWZJJ21} have theoretically proven that any loss function can be made robust to noisy labels by constraining the network output to the set of permutations over a fixed vector. 
When using a one-hot fixed vector, the output should be constrained to be one-hot as well. 
However, this constraint leads to gradients close to zero in most cases, making gradient-based optimization challenging.
To address this issue, we adopt Sparse Regularization (SR) as \citet{DBLP:conf/iccv/ZhouLWZJJ21} to enhance the performance of GCE. 
SR approximates the one-hot constraint by employing two key components: first, a network output sharpening operation to enforce a sharp output distribution, and second, $L_p$-norm ($p \textless 1$) regularization that promotes sparsity in the network output.
This straightforward approach ensures the robustness of any loss function to noisy labels without hindering the network's fitting ability.
GCE + SR only calculated on $D_n$ can be defined as:

\begin{align}
\mathcal{L}_{\mathrm{GCE+SR}} =   \sum_{x_{i,j} \in D_{n}} \frac{1-o_{i,j}^{q}}{q} + \left\|\left(\mathbf{o}_{i,j}\right)\right\|_{p}^{p}
\label{equation:loss_gce_sr}
\end{align}
where $p \leq 1$ denotes a hyperparameter. 

In this way, we can achieve better robustness to asymmetric noise caused by incomplete annotation.

\begin{table}
\centering
\footnotesize
\arrayrulecolor{black}
\begin{tabular}{llcc} 
\toprule
\multirow{2}{*}{Dataset~}  & \multicolumn{1}{l}{\multirow{2}{*}{Type}} & \multicolumn{1}{c}{Inaccurate} & \multicolumn{1}{c}{Incomplete}  \\
                           & \multicolumn{1}{c}{}                      & \multicolumn{1}{c}{Rate}       & \multicolumn{1}{c}{Rate}            \\
                           \cmidrule(lr){1-4}
\multirow{2}{*}{BC5CDR}    & Chemical                                  & 2.01                           & 36.86                               \\
                           & Disease                                   & 1.64                           & 53.27                               \\
                           \cmidrule(lr){1-4}
\multirow{4}{*}{CoNLL2003} & PER                                       & 17.64                          & 17.89                               \\
                           & LOC                                       & 0.02                           & 34.80                                \\
                           & ORG                                       & 9.53                           & 39.41                               \\
                           & MISC                                      & 0.00                              & 79.93                               \\
                           \toprule
\end{tabular}
\caption{The token-level quality of distant labels on training sets in \citet{zhou-etal-2022-distantly} settings.}
\label{tab_two}
\arrayrulecolor{black}
\end{table}

\subsection{Training}
\noindent
We weighted the losses as follows:
\begin{align}
\mathcal{L}_{\mathrm{Final}} =  \eta \mathcal{L}_{\mathrm{MFL}} +
(1-\eta)\mathcal{L}_{\mathrm{CL}} + \mathcal{L}_{\mathrm{GCE+SR}} 
\label{equation:loss_final}
\end{align}
$\eta$ is hyperparameter to control the weight between $\mathcal{L}_{\mathrm{MFL}}$ and $\mathcal{L}_{\mathrm{CL}}$ only calculated on spans labeled as entities. As introduced before, $\mathcal{L}_{\mathrm{GCE+SR}}$ is only calculated on spans labeled as non-entities.

\begin{table}
\centering
\small
\arrayrulecolor{black}
\begin{tabular}{lccc} 
\toprule
\multicolumn{1}{l}{Dataset} & Types & Train & Test \\
\cmidrule(lr){1-4}
{CoNLL2003}  & 4 & 14041 & 3453 \\
{OntoNotes5.0}  & 18 & 115812 & 12217 \\
{Webpage} & 4 & 385 & 135 \\
{BC5CDR} & 2 & 4560 & 4797\\
{EC} & 5 & 3657 & 798\\
\toprule
\end{tabular}
\caption{Statistics of five DS-NER datasets.}
\arrayrulecolor{black}
\label{tab_datasets}
\end{table}
\renewcommand\arraystretch{1.2}
\begin{table*}
\centering
\arrayrulecolor{black}
\scriptsize
\setlength{\tabcolsep}{2mm}{
\begin{tabular}{lccccccccccccccc}
\toprule
\multirow{2}{*}{Method} & \multicolumn{3}{c}{CoNLL2003} & \multicolumn{3}{c}{Webpage}& \multicolumn{3}{c}{OntoNotes5.0}& \multicolumn{3}{c}{BC5CDR}& \multicolumn{3}{c}{EC}\\ 
\cline{2-16} & \textbf{P}& \textbf{R}& \textbf{F1}& \textbf{P}& \textbf{R}& \textbf{F1}& \textbf{P}& \textbf{R} & \textbf{F1}  & \textbf{P} & \textbf{R} & \textbf{F1} & \textbf{P} & \textbf{R} & \textbf{F1}\\
\hline
KB-Matching & 81.13 & 63.75 & 71.40 & 62.59 & 45.14 & 52.45 & 63.86 & 55.71 & 59.51 & 85.90 & 48.20 & 61.70 & - & - & 44.02\\
\hline
BiLSTM-CRF & 75.50 & 49.10 & 59.50 & 58.05 & 34.59 & 43.34 & 68.44 & 64.50 & 66.41 & 83.60 & 52.40 & 64.40 & - & - & 54.59 \\
RoBERTa& 82.29 & 70.47 & 75.93 & 59.24 & 62.84 & 60.98 & 66.99 & 69.51 & 68.23 & 79.30 & 66.50 & 72.30 & - & - & - \\
\hline
LRNT$^\dag$  & 79.91 & 61.87 & 69.74 & 46.70 & 48.83 & 47.74 & 67.36 & 68.02 & 67.69 & - & - & - & - & - & - \\
Co-teaching+$^*$ & 86.04 & 68.74 & 76.42 & 61.65 & 55.41 & 58.36 & 66.63 & 69.32 & 67.95 & - & - & - & - & - & - \\
JoCoR$^*$ & 83.65 & 69.69 & 76.04 & 62.14 & 58.78 & 60.42 & 66.74 & 68.74 & 67.73 & - & - & - & - & - & - \\
NegSampling$^\dag$ & 80.17 & 77.72 & 78.93 & \underline{70.16} & 58.78 & 63.97 & 64.59 & \textbf{72.39} & 68.26 & - & - & - & - & - & 66.17 \\
NegSampling+$^\dag$ & - & - & - & - & - & - & - & - & - & - & - & - & - & - & 67.03 \\
SCDL$^*$ & \textbf{87.96} & 79.82 & 83.69 & 68.71 & \textbf{68.24} & \underline{68.47} & \underline{67.49} & 69.77 & \underline{68.61} & - & - & - & - & - & - \\
AutoNER$^\dag$ & 75.21 & 60.40 & 67.00 & 48.82 & 54.23 & 51.39 & 64.63 & 69.95 & 67.18 & \underline{79.80} & 58.60 & 67.50 & - & - & - \\
BOND$^*$ & 82.05 & 80.92 & 81.48 & 67.37 & 64.19 & 65.74 & 67.14 & 69.61 & 68.35 & 78.80 & 66.60 & 72.10 & - & - & - \\
RoSTER$^*$ & 85.90 & \underline{84.90} & \underline{85.40} & - & - & - & - & - & - & 73.30 & 72.60 & 72.90 & - & - & - \\
Conf-MPU$^\dag$ & 78.58 & 79.75 & 79.16 & - & - & - & - & - & - & 69.79 & \textbf{86.42} & \underline{77.22} & - & - & - \\
CReDEL$^*$ & - & - & - & - & - & - & - & - & - & 65.20 & \underline{80.60} & 72.10 & - & - & - \\
SCL-RAI$^\dag$ & - & - & - & - & - & - & - & - & - & - & - & - & - & - & 71.24 \\
\hline
\textbf{Ours} &\underline{86.25} & \textbf{86.95} & \textbf{86.59} & \textbf{78.40}& \underline{66.22} & \textbf{71.79} &\textbf{69.24} &\underline{70.21} &\textbf{69.72} & \textbf{81.74} & \underline{76.88} &\textbf{79.23}&\textbf{73.13} &\textbf{73.53}&\textbf{73.33}\\
\toprule
\end{tabular}
}
\arrayrulecolor{black}
\caption{Results on BC5CDR, CoNLL2003, OntoNotes5.0, Webpage. We report the baseline results from \citet{zhang-2022-improve}, \citet{zhang-etal-2021-improving-distantly}, \citet{meng-etal-2021-distantly}, and \citet{si-etal-2022-scl}. $\dag$: methods that only consider the incomplete annotation.  ${*}$: methods that consider two types of noise equally.}
\label{tab_main}
\end{table*}

\begin{table}
\small
\centering
\renewcommand{\arraystretch}{1.1}
\scriptsize
\setlength{\tabcolsep}{2mm}{
\begin{tabular}{lccc}
\toprule
\multirow{2}{*}{Method} & \multicolumn{3}{c}{Webpage}\\
\cline{2-4} & \textbf{P}& \textbf{R}& \textbf{F1} \\
\hline
{Focal Loss}&{66.67}&{55.77}&{60.74} \\
{CE}&{67.32}&{56.41}&{61.38} \\
{GCE + SR}&{67.45}&{57.17}&{61.88} \\
{CE \& GCE + SR}&{68.46}&{58.32}&{62.98}\\
{\textbf{SANTA}}&{78.40}&{66.22}&{\textbf{71.79}}\\
\hline
{w/o MFL }&{-}&{-}&{-}\\
{\quad w. GCE + SR}&{77.78}&{61.49}&{68.68}\\
{\quad w. CE}&{73.33}&{66.89}&{69.96}\\
{w/o Memory Label Smoothing in MLF}&{74.02}&{63.51}&{68.36}\\
{\quad w. Label Smoothing in MLF}&{74.05}&{65.54}&{69.53}\\
\hline
{w/o Entity-aware KNN}&{77.66}&{49.32}&{60.33}\\
{\quad w. KNN-augmented Inference}&{67.94}&{60.14}&{63.80}\\
{\quad w. Entity-aware CL}&{87.84}&{43.92}&{58.56}\\
\hline
{w/o GCE + SR} &{-}&{-}&{-}\\ 
{\quad w. GCE}&{77.52}&{66.03}&{71.32}\\
{\quad w. CE}&{76.56}&{66.22}&{71.01}\\
{\quad w. MFL}&{72.31}&{66.43}&{69.25}\\
\hline
{w/o Boundary Mixup}&{70.71}&{47.30}&{56.68}\\
{\quad w. Mixup}&{76.98}&{65.54}&{70.80}\\
\toprule
\end{tabular}
}
\caption{Ablation study on Webpage.}
\label{tab_ablation} 
\end{table}

\section{Experiment}\label{sec:experiments}
 Compared with extensive baselines, SANTA achieves significant improved performance in five datasets.
We also conduct experiments and provide analyses to justify the effectiveness of SANTA.

\subsection{Dataset}
\noindent
We conduct experiments on five benchmark datasets, including CoNLL2003 \citep{tjong-kim-sang-de-meulder-2003-introduction}, Webpage \citep{ratinov-roth-2009-design}, OntoNotes5.0 \citep{weischedel2013ontonotes}, BC5CDR \citep{li2015annotating} and EC \citep{DBLP:conf/coling/YangCLHZ18}.
For CoNLL2003, OntoNotes5.0, Webpage, \citet{DBLP:conf/kdd/LiangYJEWZZ20} re-annotates the training set by distant supervision, and uses the original development and test set.
We keep the same knowledge bases as \citet{shang-etal-2018-learning} in BC5CDR.
For EC, \citet{DBLP:conf/coling/YangCLHZ18} uses the distant supervision to get training data, and labels development and test set by crowd-sourcing.
Statistics of five datasets are shown in Table \ref{tab_datasets}.


\subsection{Evaluation Metrics and Baselines}
\noindent
To compare with baselines, we use Precision (P), Recall (R), and F1 score as the evaluation metrics.
We compare SANTA with different groups of baseline methods, including supervised methods and distantly supervised methods. 
We also present the results of \textbf{KB-Matching}, which directly uses knowledge bases to annotate the test sets.

\paragraph{Supervised Methods}
We select \textbf{BiLSTM-CRF} \citep{ma-hovy-2016-end} and \textbf{RoBERTa} \citep{liu2019roberta} as original supervised methods.
As trained on noisy text, these methods achieve poor performance on DS-NER datasets. 

\paragraph{Distantly-Supervised Methods}


We compare several DS-NER baselines, including:
(1) methods that only consider the incomplete annotation:
\textbf{Conf-MPU} \citep{zhou-etal-2022-distantly} uses multi-class PU-learning loss to better estimate the loss.
\textbf{AutoNER} \citep{shang-etal-2018-learning} modifies the standard CRF to get better performance under the noise. 
\textbf{LRNT} \citep{DBLP:conf/emnlp/CaoHCLJ19} tries to reduce the negative effect of noisy labels, leaving training data unexplored fully.
\textbf{NegSampling} \citep{DBLP:conf/iclr/LiL021} and \textbf{Weighted NegSampling} \citep{li-etal-2022-rethinking} uses down-sampling in non-entities to relief the misleading from incomplete annotation.
\textbf{SCL-RAI} \citep{si-etal-2022-scl} uses span-based supervised contrastive-learning loss and designed inference method to improve the robustness against incomplete annotation.
(2) methods that consider two types of noise equally:
\textbf{Co-teaching+} \citep{DBLP:conf/icml/Yu0YNTS19} and \textbf{JoCoR} \citep{DBLP:conf/cvpr/WeiFC020} are two classical methods to handle noisy labels in computer vision area.
\textbf{BOND} \citep{DBLP:conf/kdd/LiangYJEWZZ20} and \textbf{SCDL} \citep{zhang-etal-2021-improving} proposes teacher-student network to reduce the noise from distant labels.
\textbf{RoSTER} \citep{meng-etal-2021-distantly} adopts GCE loss, self-training and noisy label removal step to improve the robustness.
\textbf{CReDEL} \citep{ying-etal-2022-label} proposes an automatic distant label refinement model via contrastive-learning as a plug-in module.

\subsection{Experimental Settings}
For a fair comparison, (1) we use BERT-base \citep{devlin-etal-2019-bert} as the encoder the same as \citet{si-etal-2022-scl} and \citet{zhang-2022-improve} for CoNLL2003, OntoNotes5.0, Webpage and EC; (2) for BC5CDR in the biomedical domain, we use BioBERT-base \citep{DBLP:journals/bioinformatics/LeeYKKKSK20} the same as \citet{zhang-2022-improve}.
We use Adam \citep{DBLP:journals/corr/KingmaB14} as our optimizer.
We list detailed hyperparameters in Table \ref{tab_hyper}. 
Experiments are run on NVIDIA-P40 and NVIDIA-A100.

\subsection{Main Results}
\label{sec:main_results}
\noindent
Table \ref{tab_main} presents the main results of SANTA.
From these results, the following four insights can be drawn.
(1) on all five datasets, SANTA achieves the best F1 performance among all DS-NER baselines, and strikes a good balance between precision and recall, demonstrating superiority when trained on distantly-supervised text;
(2) compared to original supervised methods, including BiLSTM-CRF and RoBERTa, SANTA improves the F1 score with an average increase of 18.48\% and 7.47\% respectively, which demonstrates the necessity of DS-NER models and the effectiveness;
(3) compared with methods only focusing on incomplete annotation such as NegSampling, SANTA achieves more balanced precision and recall, showing the necessity to handle two types of noise.
(4) compared with methods handling two types of noise with the same strategy such as RoSTER, SANTA achieves better performance both in precision and recall, demonstrating the effectiveness of separate handling.

\begin{table*}
\small
\centering
\renewcommand{\arraystretch}{1.05}
\begin{tabular}{l}
\toprule
\textbf{Distant Match}:  \textcolor[rgb]{1,0,0}{[Johnson]$_{\mathrm{PER}}$} is the second manager to be hospitalized after California \textcolor[rgb]{1,0,0}{[Angels]$_{\mathrm{PER}}$} \\skipper \textcolor[rgb]{1,0,0}{[John]$_{\mathrm{PER}}$} McNamara was admitted to New \textcolor[rgb]{1,0,0}{[York]$_{\mathrm{PER}}$} ’s \textcolor[rgb]{1,0,0}{[Columbia]$_{\mathrm{PER}}$} Presby Hospital .\\
\textbf{Ground Truth}:  \textcolor[rgb]{1,0,0}{[Johnson]$_{\mathrm{PER}}$} is the second manager to be hospitalized after \textcolor[rgb]{0,0,0.7}{[California Angels]$_{\mathrm{ORG}}$} \\skipper \textcolor[rgb]{1,0,0}{[John McNamara]$_{\mathrm{PER}}$} was admitted to \textcolor[rgb]{0,0.7,0}{[New York]$_{\mathrm{LOC}}$} ’s \textcolor[rgb]{0,0,0.7}{[Columbia Presby Hospital]$_{\mathrm{ORG}}$} .\\
\hline
\textbf{Conf-MPU}: \textcolor[rgb]{1,0,0}{[Johnson]$_{\mathrm{PER}}$} is the second manager to be hospitalized after \underline{\textcolor[rgb]{0,0.7,0}{[California]$_{\mathrm{LOC}}$} \textcolor[rgb]{1,0,0}{[Angels]$_{\mathrm{PER}}$}} \\skipper \textcolor[rgb]{1,0,0}{[John McNamara]$_{\mathrm{PER}}$} was admitted to \textcolor[rgb]{0,0.7,0}{[New York]$_{\mathrm{LOC}}$} ’s \underline{\textcolor[rgb]{1,0,0}{[Columbia]$_{\mathrm{PER}}$} Presby Hospital} .\\
\textbf{SCDL}: \textcolor[rgb]{1,0,0}{[Johnson]$_{\mathrm{PER}}$} is the second manager to be hospitalized after \underline{\textcolor[rgb]{0,0.7,0}{[California]$_{\mathrm{LOC}}$} \textcolor[rgb]{1,0,0}{[Angels]$_{\mathrm{PER}}$}} \\skipper \textcolor[rgb]{1,0,0}{[John McNamara]$_{\mathrm{PER}}$} was admitted to \textcolor[rgb]{0,0.7,0}{[New York]$_{\mathrm{LOC}}$} ’s \textcolor[rgb]{0,0,0.7}{[Columbia Presby Hospital]$_{\mathrm{ORG}}$} .\\
\textbf{Ours}: \textcolor[rgb]{1,0,0}{[Johnson]$_{\mathrm{PER}}$} is the second manager to be hospitalized after \textcolor[rgb]{0,0,0.7}{[California Angels]$_{\mathrm{ORG}}$} \\skipper \textcolor[rgb]{1,0,0}{[John McNamara]$_{\mathrm{PER}}$} was admitted to \textcolor[rgb]{0,0.7,0}{[New York]$_{\mathrm{LOC}}$} ’s \textcolor[rgb]{0,0,0.7}{[Columbia Presby Hospital]$_{\mathrm{ORG}}$} .\\
\toprule
\end{tabular}
\caption{Case study with SANTA and baselines.}
\label{tab_case} 
\end{table*}

\subsection{Analysis}

\paragraph{Ablation Study.}
To further validate the effectiveness of each component, we compare SANTA with the fine-grained ablations by removing one component at a time in Table \ref{tab_ablation}.
(1) MFL improves the precision and recall compared with using noise-tolerant GCE + SR, indicating that MFL can help model to avoid underfitting. When using CE instead of MFL, percision is significantly reduced showing MFL can better handle entity ambiguity problem. Meanwhile, designed MLS can reduce the fluctuation of Focal Loss to achieve better performance, and better than static Label Smoothing;
(2) Entity-aware KNN consists of KNN-augmented Inference and Entity-aware CL, the significantly reduction of F1 score and only using one of them achieves poor performance, both showing the design of Entity-aware KNN is necessary;
(3) compared with GCE, CE and MFL, GCE + SR shows the strong capability to handle the incomplete annotation noise, due to the improvement of robustness and asymmetric noise condition;
(4) Boundary Mixup can significantly improve the recall, which indicates it can help model to make more correct prediction of the samples around the decision boundary and alleviate the decision boundary shifting problem. Boundary Mixup reduces randomness of sampling in Mixup and focus on examples around the decision boundary to better handle the incomplete annotation in DS-NER.

\begin{table}
\footnotesize
\centering
\renewcommand{\arraystretch}{1.05}
\begin{tabular}{l}
\toprule
\textbf{Distant Match}: \textcolor[rgb]{1,0,0}{[Arafat]$_{\mathrm{PER}}$} to meet \textcolor[rgb]{1,0,0}{[Peres]$_{\mathrm{PER}}$} in \\ \textcolor[rgb]{1,0,0}{[Gaza]$_{\mathrm{PER}}$} after flight ban.\\
\textbf{Ground Truth}:  \textcolor[rgb]{1,0,0}{[Arafat]$_{\mathrm{PER}}$} to meet \textcolor[rgb]{1,0,0}{[Peres]$_{\mathrm{PER}}$} in \\\textcolor[rgb]{0,0.7,0}{[Gaza]$_{\mathrm{LOC}}$} after flight ban.\\

\textbf{Span-based NER Model}: \textcolor[rgb]{1,0,0}{[Arafat]$_{\mathrm{PER}}$} to meet \\ \textcolor[rgb]{1,0,0} {[Peres]$_{\mathrm{PER}}$} in  \textcolor[rgb]{1,0,0}{[Gaza]$_{\mathrm{PER: 0.42}}$} after flight ban.\\
\textbf{w. Entity-aware KNN}: \textcolor[rgb]{1,0,0}{[Arafat]$_{\mathrm{PER}}$} to meet \textcolor[rgb]{1,0,0}{[Peres]$_{\mathrm{PER}}$} \\in \textcolor[rgb]{0,0.7,0}{[Gaza]$_{\mathrm{LOC: 0.46}}$} after flight ban.\\
\hline

\textbf{Distant Match}: \textcolor[rgb]{0,0,0.7}{[EC]$_{\mathrm{ORG}}$} rejects German call to \\ boycott British lamb.\\
\textbf{Ground Truth}: \textcolor[rgb]{0,0,0.7}{[EC]$_{\mathrm{ORG}}$} rejects \textcolor[HTML]{EC9704}{[German]$_{\mathrm{MISC}}$} call \\ to boycott \textcolor[HTML]{EC9704}{[British]$_{\mathrm{MISC}}$} lamb.\\

\textbf{Span-based NER Model}: \textcolor[rgb]{0,0,0.7}{[EC]$_{\mathrm{ORG}}$} rejects \\ \textcolor[HTML]{EC9704}{[German]$_{\mathrm{MISC}}$} call to boycott {[British]\textcolor[HTML]{EC9704}{$_{\mathrm{MISC: 0.31}}$}} lamb.\\
\textbf{w. Boundary Mixup}: \textcolor[rgb]{0,0,0.7}{[EC]$_{\mathrm{ORG}}$} rejects \textcolor[HTML]{EC9704}{[German]$_{\mathrm{MISC}}$} \\call  to boycott \textcolor[HTML]{EC9704}{[British]$_{\mathrm{MISC: 0.56}}$} lamb.\\

\toprule

\end{tabular}
\caption{Exploring Entity-aware KNN \& Boundary Mixup.
We give the selected scores from SANTA. }
\label{explore} 
\end{table}

\paragraph{Case Study.}
We also perform case study to better understand the advantage of SANTA in Table \ref{tab_case}.
We show the prediction of Conf-MPU only focusing on incomplete annotation noise, and SCDL handling two types of noise with same strategy.
Conf-MPU is able to learn from labeled DS-NER text and slightly learn to generalize, such like ``Johnson" and ``John McNamara" can be recognized correctly.
But the limited generalization capability leads to memorize ``Columbia" and ``Angels" with wrong labels.
Meanwhile, Conf-MPU can not handle entity ambiguity well because it ignores the inaccurate annotation noise, such as wrongly predicting the type of span ``California Angels" and ``Columbia Presby Hospital".
SCDL is able to generalize better and slightly handle entity ambiguity, due to the comprehensive consideration of two types of noise.
But it is still impacted by entity ambiguity problem in difficult span ``California Angels" .
SANTA can further detect the noisy labels via separate strategies and correctly recognize all the entities in this sentence from CoNLL2003 training set.

\paragraph{Strategies Exploration.}
We further explore our strategies as following:
(1) as shown in Table \ref{tab_ablation}, we simply use Focal Loss, CE, GCE + SR in Span-based NER model as baselines. Meanwhile, we use CE \& GCE + SR separately for the spans labeled as entities suffered from inaccurate annotation \& the spans labeled as non-entities suffered from incomplete annotation.
CE \& GCE + SR and SANTA achieve better performance, showing that our motivation of separately handing is helpful and well-designed separate strategies is powerful.
(2) as shown cases in Table \ref{explore}, we can observe that only use Entity-aware KNN and only use Boundary Mixup can both help model to make more accurate predictions.
Entity-aware KNN augments the score from model and finally get the right label prediction, alleviating the entity ambiguity problem.
Boundary Mixup correctly help model to recall span ``British", which indicates decision boundary shifting problem is relieved.

\begin{table}
\scriptsize
\setlength\tabcolsep{1.pt}
    \centering
        \begin{tabular}{lccccc}
    \toprule
    Name & BC5CDR & CoNLL2003 & OntoNotes5.0 & Webpage &EC\\
    \midrule
    Learning Rate &{1e-5} &{1e-5} &{1e-5} &{1e-5} &{1e-5}  \\
    Batch Size &{16} &{16} &{16} &{12} &{16} \\
    dim of $W$ in \ref{W_dim} &{256} &{256} &{256} &{128} &{256} \\
    $K$ in Entity-aware KNN &{64} &{64} &{64} &{16} &{64}\\
    $G$ in eq. \ref{equation:loss_mfl_pre}&{1} &{3} &{1} &{1} &{1}\\
    $\epsilon$ in Boundary Mixup &{0.5} &{0.5} &{0.5} &{0.5} &{0.5}\\
    $\lambda$ in eq. \ref{equation:loss_mfl_pre} &{0.8} &{0.8} &{0.8} &{0.8} &{0.8}\\
    $\alpha$ in eq. \ref{equation:loss_mfl} &{0.5} &{0.5} &{0.5} &{0.5} &{0.5}  \\
    $\gamma$ in eq. \ref{equation:loss_mfl}  &{2} &{2} &{2} &{2} &{2}  \\
    $\tau$ in eq. \ref{equation:loss_cl}&{0.05} &{0.05} &{0.05} &{0.05} &{0.05}  \\
    $\mu$ in eq. \ref{equation:loss_knn} &{0.7} &{0.3} &{0.3} &{0.7} &{0.7}  \\
    $p$ in eq. \ref{equation:loss_gce_sr} &{0.5} &{0.5} &{0.5} &{0.5} &{0.5} \\
    $q$ in eq. \ref{equation:loss_gce_sr}&{0.3} &{0.3} &{0.3} &{0.3} &{0.3}   \\
    $\alpha'$ in eq. \ref{equation:loss_mixup} &{0.2} &{0.2} &{0.2} &{0.2} &{0.2}  \\
    $\eta$ in eq. \ref{equation:loss_final} &{0.9} &{0.9} &{0.9} &{0.9} &{0.9} \\
    \bottomrule
    \end{tabular}
    \caption{Hyperparameters}
    \label{tab_hyper}
\vspace{-0.1in}
\end{table}

\section{Conclusion}
Inaccurate and incomplete annotation noise are two types
of noise in DS-NER. 
We propose SANTA to use separate strategies for two types of noise.
For inaccurate annotation, we propose Memory-smoothed Focal Loss and Entity-aware KNN to relief the ambiguity problem. 
For incomplete annotation, we utilize noise-robust loss GCE + SR and propose Boundary Mixup to improve the robustness and mitigate the decision boundary shifting problem.
Experiments show that SANTA achieves state-of-the-art methods on five DS-NER datasets and the separate strategies are effective.

\section*{Limitations}
Our proposed work is dedicated to considering the noise in DS-NER, and our noise-specific analyses are all based on this task.
Therefore, if it were not for DS-NER task, our model would not necessarily be robust compared to other task-specific methods.
Also, our approach is based entirely on previous experimental settings in DS-NER, so we do not consider how to reduce noise from the distant supervision process, e.g., designing models to help the annotation process rather than learning to reduce noise from the distantly-supervised text.
Designing models to help the distant supervision process could be a direction for future study.

\section*{Acknowledgements}
We thank all the reviewers for their comments.
This paper is supported by the National Key R\&D Program of China under Grand No.2018AAA0102003, the National Science Foundation of China under Grant No.61936012. 
Our code can be found in \href{https://github.com/PKUnlp-icler/SANTA}{https://github.com/PKUnlp-icler/SANTA}.

\bibliography{anthology,custom}

\begin{thebibliography}{39}
\expandafter\ifx\csname natexlab\endcsname\relax\def\natexlab#1{#1}\fi

\bibitem[{Cao et~al.(2019)Cao, Hu, Chua, Liu, and Ji}]{DBLP:conf/emnlp/CaoHCLJ19}
Yixin Cao, Zikun Hu, Tat{-}Seng Chua, Zhiyuan Liu, and Heng Ji. 2019.
\newblock \href {https://doi.org/10.18653/v1/D19-1025} {Low-resource name tagging learned with weakly labeled data}.
\newblock In \emph{Proceedings of the 2019 Conference on Empirical Methods in Natural Language Processing and the 9th International Joint Conference on Natural Language Processing, {EMNLP-IJCNLP} 2019, Hong Kong, China, November 3-7, 2019}, pages 261--270. Association for Computational Linguistics.

\bibitem[{Charoenphakdee et~al.(2019)Charoenphakdee, Lee, and Sugiyama}]{DBLP:conf/icml/CharoenphakdeeL19}
Nontawat Charoenphakdee, Jongyeong Lee, and Masashi Sugiyama. 2019.
\newblock \href {http://proceedings.mlr.press/v97/charoenphakdee19a.html} {On symmetric losses for learning from corrupted labels}.
\newblock In \emph{Proceedings of the 36th International Conference on Machine Learning, {ICML} 2019, 9-15 June 2019, Long Beach, California, {USA}}, volume~97 of \emph{Proceedings of Machine Learning Research}, pages 961--970. {PMLR}.

\bibitem[{Devlin et~al.(2019)Devlin, Chang, Lee, and Toutanova}]{devlin-etal-2019-bert}
Jacob Devlin, Ming-Wei Chang, Kenton Lee, and Kristina Toutanova. 2019.
\newblock \href {https://doi.org/10.18653/v1/N19-1423} {{BERT}: Pre-training of deep bidirectional transformers for language understanding}.
\newblock In \emph{Proceedings of the 2019 Conference of the North {A}merican Chapter of the Association for Computational Linguistics: Human Language Technologies, Volume 1 (Long and Short Papers)}, pages 4171--4186, Minneapolis, Minnesota. Association for Computational Linguistics.

\bibitem[{Ghosh et~al.(2017)Ghosh, Kumar, and Sastry}]{DBLP:conf/aaai/GhoshKS17}
Aritra Ghosh, Himanshu Kumar, and P.~S. Sastry. 2017.
\newblock \href {http://aaai.org/ocs/index.php/AAAI/AAAI17/paper/view/14759} {Robust loss functions under label noise for deep neural networks}.
\newblock In \emph{Proceedings of the Thirty-First {AAAI} Conference on Artificial Intelligence, February 4-9, 2017, San Francisco, California, {USA}}, pages 1919--1925. {AAAI} Press.

\bibitem[{Jie et~al.(2019)Jie, Xie, Lu, Ding, and Li}]{jie-etal-2019-better}
Zhanming Jie, Pengjun Xie, Wei Lu, Ruixue Ding, and Linlin Li. 2019.
\newblock \href {https://doi.org/10.18653/v1/N19-1079} {Better modeling of incomplete annotations for named entity recognition}.
\newblock In \emph{Proceedings of the 2019 Conference of the North {A}merican Chapter of the Association for Computational Linguistics: Human Language Technologies, Volume 1 (Long and Short Papers)}, pages 729--734, Minneapolis, Minnesota. Association for Computational Linguistics.

\bibitem[{Kingma and Ba(2015)}]{DBLP:journals/corr/KingmaB14}
Diederik~P. Kingma and Jimmy Ba. 2015.
\newblock \href {http://arxiv.org/abs/1412.6980} {Adam: {A} method for stochastic optimization}.
\newblock In \emph{3rd International Conference on Learning Representations, {ICLR} 2015, San Diego, CA, USA, May 7-9, 2015, Conference Track Proceedings}.

\bibitem[{Lee et~al.(2020)Lee, Yoon, Kim, Kim, Kim, So, and Kang}]{DBLP:journals/bioinformatics/LeeYKKKSK20}
Jinhyuk Lee, Wonjin Yoon, Sungdong Kim, Donghyeon Kim, Sunkyu Kim, Chan~Ho So, and Jaewoo Kang. 2020.
\newblock \href {https://doi.org/10.1093/bioinformatics/btz682} {Biobert: a pre-trained biomedical language representation model for biomedical text mining}.
\newblock \emph{Bioinform.}, 36(4):1234--1240.

\bibitem[{Li et~al.(2015)Li, Sun, Johnson, Sciaky, Wei, Leaman, Davis, Mattingly, Wiegers, and Lu}]{li2015annotating}
Jiao Li, Yueping Sun, R~Johnson, Daniela Sciaky, Chih-Hsuan Wei, Robert Leaman, Allan~Peter Davis, Carolyn~J Mattingly, Thomas~C Wiegers, and Zhiyong Lu. 2015.
\newblock Annotating chemicals, diseases, and their interactions in biomedical literature.
\newblock In \emph{Proceedings of the fifth BioCreative challenge evaluation workshop}, pages 173--182. The Fifth BioCreative Organizing Committee.

\bibitem[{Li et~al.(2020)Li, Liu, and Shi}]{li2020empirical}
Yangming Li, Lemao Liu, and Shuming Shi. 2020.
\newblock Empirical analysis of unlabeled entity problem in named entity recognition.
\newblock \emph{arXiv preprint arXiv:2012.05426}.

\bibitem[{Li et~al.(2021)Li, Liu, and Shi}]{DBLP:conf/iclr/LiL021}
Yangming Li, Lemao Liu, and Shuming Shi. 2021.
\newblock \href {https://openreview.net/forum?id=5jRVa89sZk} {Empirical analysis of unlabeled entity problem in named entity recognition}.
\newblock In \emph{9th International Conference on Learning Representations, {ICLR} 2021, Virtual Event, Austria, May 3-7, 2021}. OpenReview.net.

\bibitem[{Li et~al.(2022{\natexlab{a}})Li, Liu, and Shi}]{li-etal-2022-rethinking}
Yangming Li, Lemao Liu, and Shuming Shi. 2022{\natexlab{a}}.
\newblock \href {https://doi.org/10.18653/v1/2022.acl-long.497} {Rethinking negative sampling for handling missing entity annotations}.
\newblock In \emph{Proceedings of the 60th Annual Meeting of the Association for Computational Linguistics (Volume 1: Long Papers)}, pages 7188--7197, Dublin, Ireland. Association for Computational Linguistics.

\bibitem[{Li et~al.(2022{\natexlab{b}})Li, Liu, and Shi}]{li2022rethinking}
Yangming Li, Lemao Liu, and Shuming Shi. 2022{\natexlab{b}}.
\newblock Rethinking negative sampling for handling missing entity annotations.
\newblock In \emph{Proceedings of the 60th Annual Meeting of the Association for Computational Linguistics (Volume 1: Long Papers)}, pages 7188--7197.

\bibitem[{Li et~al.(2022{\natexlab{c}})Li, Liu, Yang, and Li}]{li-etal-2022-self}
Yunshui Li, Junhao Liu, Min Yang, and Chengming Li. 2022{\natexlab{c}}.
\newblock \href {https://aclanthology.org/2022.findings-emnlp.149} {Self-distillation with meta learning for knowledge graph completion}.
\newblock In \emph{Findings of the Association for Computational Linguistics: EMNLP 2022}, pages 2048--2054, Abu Dhabi, United Arab Emirates. Association for Computational Linguistics.

\bibitem[{Liang et~al.(2020)Liang, Yu, Jiang, Er, Wang, Zhao, and Zhang}]{DBLP:conf/kdd/LiangYJEWZZ20}
Chen Liang, Yue Yu, Haoming Jiang, Siawpeng Er, Ruijia Wang, Tuo Zhao, and Chao Zhang. 2020.
\newblock \href {https://doi.org/10.1145/3394486.3403149} {{BOND:} bert-assisted open-domain named entity recognition with distant supervision}.
\newblock In \emph{{KDD} '20: The 26th {ACM} {SIGKDD} Conference on Knowledge Discovery and Data Mining, Virtual Event, CA, USA, August 23-27, 2020}, pages 1054--1064. {ACM}.

\bibitem[{Lin et~al.(2020)Lin, Goyal, Girshick, He, and Doll{\'{a}}r}]{DBLP:journals/pami/LinGGHD20}
Tsung{-}Yi Lin, Priya Goyal, Ross~B. Girshick, Kaiming He, and Piotr Doll{\'{a}}r. 2020.
\newblock \href {https://doi.org/10.1109/TPAMI.2018.2858826} {Focal loss for dense object detection}.
\newblock \emph{{IEEE} Trans. Pattern Anal. Mach. Intell.}, 42(2):318--327.

\bibitem[{Liu et~al.(2019)Liu, Ott, Goyal, Du, Joshi, Chen, Levy, Lewis, Zettlemoyer, and Stoyanov}]{liu2019roberta}
Yinhan Liu, Myle Ott, Naman Goyal, Jingfei Du, Mandar Joshi, Danqi Chen, Omer Levy, Mike Lewis, Luke Zettlemoyer, and Veselin Stoyanov. 2019.
\newblock Roberta: A robustly optimized bert pretraining approach.
\newblock \emph{arXiv preprint arXiv:1907.11692}.

\bibitem[{Ma and Hovy(2016)}]{ma-hovy-2016-end}
Xuezhe Ma and Eduard Hovy. 2016.
\newblock \href {https://doi.org/10.18653/v1/P16-1101} {End-to-end sequence labeling via bi-directional {LSTM}-{CNN}s-{CRF}}.
\newblock In \emph{Proceedings of the 54th Annual Meeting of the Association for Computational Linguistics (Volume 1: Long Papers)}, pages 1064--1074, Berlin, Germany. Association for Computational Linguistics.

\bibitem[{Meng et~al.(2021)Meng, Zhang, Huang, Wang, Zhang, Ji, and Han}]{meng-etal-2021-distantly}
Yu~Meng, Yunyi Zhang, Jiaxin Huang, Xuan Wang, Yu~Zhang, Heng Ji, and Jiawei Han. 2021.
\newblock \href {https://doi.org/10.18653/v1/2021.emnlp-main.810} {Distantly-supervised named entity recognition with noise-robust learning and language model augmented self-training}.
\newblock In \emph{Proceedings of the 2021 Conference on Empirical Methods in Natural Language Processing}, pages 10367--10378, Online and Punta Cana, Dominican Republic. Association for Computational Linguistics.

\bibitem[{Peng et~al.(2022)Peng, Liu, Xie, Xu, Wang, and Peng}]{peng-etal-2022-smile}
Miao Peng, Ben Liu, Qianqian Xie, Wenjie Xu, Hua Wang, and Min Peng. 2022.
\newblock \href {https://aclanthology.org/2022.findings-emnlp.307} {{SM}i{LE}: Schema-augmented multi-level contrastive learning for knowledge graph link prediction}.
\newblock In \emph{Findings of the Association for Computational Linguistics: EMNLP 2022}, pages 4165--4177, Abu Dhabi, United Arab Emirates. Association for Computational Linguistics.

\bibitem[{Peng et~al.(2019)Peng, Xing, Zhang, Fu, and Huang}]{DBLP:conf/acl/PengXZFH19}
Minlong Peng, Xiaoyu Xing, Qi~Zhang, Jinlan Fu, and Xuanjing Huang. 2019.
\newblock \href {https://doi.org/10.18653/v1/p19-1231} {Distantly supervised named entity recognition using positive-unlabeled learning}.
\newblock In \emph{Proceedings of the 57th Conference of the Association for Computational Linguistics, {ACL} 2019, Florence, Italy, July 28- August 2, 2019, Volume 1: Long Papers}, pages 2409--2419. Association for Computational Linguistics.

\bibitem[{Ratinov and Roth(2009)}]{ratinov-roth-2009-design}
Lev Ratinov and Dan Roth. 2009.
\newblock \href {https://aclanthology.org/W09-1119} {Design challenges and misconceptions in named entity recognition}.
\newblock In \emph{Proceedings of the Thirteenth Conference on Computational Natural Language Learning ({C}o{NLL}-2009)}, pages 147--155, Boulder, Colorado. Association for Computational Linguistics.

\bibitem[{Shang et~al.(2018)Shang, Liu, Gu, Ren, Ren, and Han}]{shang-etal-2018-learning}
Jingbo Shang, Liyuan Liu, Xiaotao Gu, Xiang Ren, Teng Ren, and Jiawei Han. 2018.
\newblock \href {https://doi.org/10.18653/v1/D18-1230} {Learning named entity tagger using domain-specific dictionary}.
\newblock In \emph{Proceedings of the 2018 Conference on Empirical Methods in Natural Language Processing}, pages 2054--2064, Brussels, Belgium. Association for Computational Linguistics.

\bibitem[{Si et~al.(2022)Si, Zeng, Lin, and Chang}]{si-etal-2022-scl}
Shuzheng Si, Shuang Zeng, Jiaxing Lin, and Baobao Chang. 2022.
\newblock \href {https://aclanthology.org/2022.coling-1.202} {{SCL}-{RAI}: Span-based contrastive learning with retrieval augmented inference for unlabeled entity problem in {NER}}.
\newblock In \emph{Proceedings of the 29th International Conference on Computational Linguistics}, pages 2313--2318, Gyeongju, Republic of Korea. International Committee on Computational Linguistics.

\bibitem[{Tjong Kim~Sang and De~Meulder(2003)}]{tjong-kim-sang-de-meulder-2003-introduction}
Erik~F. Tjong Kim~Sang and Fien De~Meulder. 2003.
\newblock \href {https://aclanthology.org/W03-0419} {Introduction to the {C}o{NLL}-2003 shared task: Language-independent named entity recognition}.
\newblock In \emph{Proceedings of the Seventh Conference on Natural Language Learning at {HLT}-{NAACL} 2003}, pages 142--147.

\bibitem[{Wang et~al.(2022)Wang, Song, Liu, Lin, Cao, Li, and Sui}]{wang-etal-2022-learning-robust}
Peiyi Wang, Yifan Song, Tianyu Liu, Binghuai Lin, Yunbo Cao, Sujian Li, and Zhifang Sui. 2022.
\newblock \href {https://aclanthology.org/2022.emnlp-main.420} {Learning robust representations for continual relation extraction via adversarial class augmentation}.
\newblock In \emph{Proceedings of the 2022 Conference on Empirical Methods in Natural Language Processing}, pages 6264--6278, Abu Dhabi, United Arab Emirates. Association for Computational Linguistics.

\bibitem[{Wei et~al.(2020)Wei, Feng, Chen, and An}]{DBLP:conf/cvpr/WeiFC020}
Hongxin Wei, Lei Feng, Xiangyu Chen, and Bo~An. 2020.
\newblock \href {https://doi.org/10.1109/CVPR42600.2020.01374} {Combating noisy labels by agreement: {A} joint training method with co-regularization}.
\newblock In \emph{2020 {IEEE/CVF} Conference on Computer Vision and Pattern Recognition, {CVPR} 2020, Seattle, WA, USA, June 13-19, 2020}, pages 13723--13732. Computer Vision Foundation / {IEEE}.

\bibitem[{Weischedel et~al.(2013)Weischedel, Palmer, Marcus, Hovy, Pradhan, Ramshaw, Xue, Taylor, Kaufman, Franchini et~al.}]{weischedel2013ontonotes}
Ralph Weischedel, Martha Palmer, Mitchell Marcus, Eduard Hovy, Sameer Pradhan, Lance Ramshaw, Nianwen Xue, Ann Taylor, Jeff Kaufman, Michelle Franchini, et~al. 2013.
\newblock Ontonotes release 5.0 ldc2013t19.
\newblock \emph{Linguistic Data Consortium, Philadelphia, PA}, 23.

\bibitem[{Yang et~al.(2018)Yang, Chen, Li, He, and Zhang}]{DBLP:conf/coling/YangCLHZ18}
YaoSheng Yang, Wenliang Chen, Zhenghua Li, Zhengqiu He, and Min Zhang. 2018.
\newblock \href {https://aclanthology.org/C18-1183/} {Distantly supervised {NER} with partial annotation learning and reinforcement learning}.
\newblock In \emph{Proceedings of the 27th International Conference on Computational Linguistics, {COLING} 2018, Santa Fe, New Mexico, USA, August 20-26, 2018}, pages 2159--2169. Association for Computational Linguistics.

\bibitem[{Ying et~al.(2022)Ying, Luo, Dang, and Yu}]{ying-etal-2022-label}
Huaiyuan Ying, Shengxuan Luo, Tiantian Dang, and Sheng Yu. 2022.
\newblock \href {https://doi.org/10.18653/v1/2022.findings-naacl.203} {Label refinement via contrastive learning for distantly-supervised named entity recognition}.
\newblock In \emph{Findings of the Association for Computational Linguistics: NAACL 2022}, pages 2656--2666, Seattle, United States. Association for Computational Linguistics.

\bibitem[{Yu et~al.(2019)Yu, Han, Yao, Niu, Tsang, and Sugiyama}]{DBLP:conf/icml/Yu0YNTS19}
Xingrui Yu, Bo~Han, Jiangchao Yao, Gang Niu, Ivor~W. Tsang, and Masashi Sugiyama. 2019.
\newblock \href {http://proceedings.mlr.press/v97/yu19b.html} {How does disagreement help generalization against label corruption?}
\newblock In \emph{Proceedings of the 36th International Conference on Machine Learning, {ICML} 2019, 9-15 June 2019, Long Beach, California, {USA}}, volume~97 of \emph{Proceedings of Machine Learning Research}, pages 7164--7173. {PMLR}.

\bibitem[{Zeng et~al.(2021)Zeng, Wu, and Chang}]{zeng-etal-2021-sire}
Shuang Zeng, Yuting Wu, and Baobao Chang. 2021.
\newblock \href {https://doi.org/10.18653/v1/2021.findings-acl.47} {{SIRE}: Separate intra- and inter-sentential reasoning for document-level relation extraction}.
\newblock In \emph{Findings of the Association for Computational Linguistics: ACL-IJCNLP 2021}, pages 524--534, Online. Association for Computational Linguistics.

\bibitem[{Zhang(2022)}]{zhang-2022-improve}
Bryan Zhang. 2022.
\newblock \href {https://aclanthology.org/2022.amta-upg.9} {Improve {MT} for search with selected translation memory using search signals}.
\newblock In \emph{Proceedings of the 15th Biennial Conference of the Association for Machine Translation in the Americas (Volume 2: Users and Providers Track and Government Track)}, pages 123--131, Orlando, USA. Association for Machine Translation in the Americas.

\bibitem[{Zhang et~al.(2018)Zhang, Ciss{\'{e}}, Dauphin, and Lopez{-}Paz}]{DBLP:conf/iclr/ZhangCDL18}
Hongyi Zhang, Moustapha Ciss{\'{e}}, Yann~N. Dauphin, and David Lopez{-}Paz. 2018.
\newblock \href {https://openreview.net/forum?id=r1Ddp1-Rb} {mixup: Beyond empirical risk minimization}.
\newblock In \emph{6th International Conference on Learning Representations, {ICLR} 2018, Vancouver, BC, Canada, April 30 - May 3, 2018, Conference Track Proceedings}. OpenReview.net.

\bibitem[{Zhang et~al.(2021{\natexlab{a}})Zhang, Yu, Liu, Zhang, Sheng, Mengge, and Xu}]{zhang-etal-2021-improving-distantly}
Xinghua Zhang, Bowen Yu, Tingwen Liu, Zhenyu Zhang, Jiawei Sheng, Xue Mengge, and Hongbo Xu. 2021{\natexlab{a}}.
\newblock \href {https://doi.org/10.18653/v1/2021.findings-emnlp.131} {Improving distantly-supervised named entity recognition with self-collaborative denoising learning}.
\newblock In \emph{Findings of the Association for Computational Linguistics: EMNLP 2021}, pages 1518--1529, Punta Cana, Dominican Republic. Association for Computational Linguistics.

\bibitem[{Zhang et~al.(2021{\natexlab{b}})Zhang, Yu, Shu, Mengge, Liu, and Guo}]{zhang-etal-2021-improving}
Zhenyu Zhang, Bowen Yu, Xiaobo Shu, Xue Mengge, Tingwen Liu, and Li~Guo. 2021{\natexlab{b}}.
\newblock \href {https://doi.org/10.18653/v1/2021.findings-acl.8} {From what to why: Improving relation extraction with rationale graph}.
\newblock In \emph{Findings of the Association for Computational Linguistics: ACL-IJCNLP 2021}, pages 86--95, Online. Association for Computational Linguistics.

\bibitem[{Zhang and Sabuncu(2018)}]{DBLP:conf/nips/ZhangS18}
Zhilu Zhang and Mert~R. Sabuncu. 2018.
\newblock \href {https://proceedings.neurips.cc/paper/2018/hash/f2925f97bc13ad2852a7a551802feea0-Abstract.html} {Generalized cross entropy loss for training deep neural networks with noisy labels}.
\newblock In \emph{Advances in Neural Information Processing Systems 31: Annual Conference on Neural Information Processing Systems 2018, NeurIPS 2018, December 3-8, 2018, Montr{\'{e}}al, Canada}, pages 8792--8802.

\bibitem[{Zhou et~al.(2022{\natexlab{a}})Zhou, Liu, and Tu}]{zhou-etal-2022-improving}
Hao Zhou, Gongshen Liu, and Kewei Tu. 2022{\natexlab{a}}.
\newblock \href {https://doi.org/10.18653/v1/2022.naacl-main.121} {Improving constituent representation with hypertree neural networks}.
\newblock In \emph{Proceedings of the 2022 Conference of the North American Chapter of the Association for Computational Linguistics: Human Language Technologies}, pages 1682--1692, Seattle, United States. Association for Computational Linguistics.

\bibitem[{Zhou et~al.(2022{\natexlab{b}})Zhou, Li, and Li}]{zhou-etal-2022-distantly}
Kang Zhou, Yuepei Li, and Qi~Li. 2022{\natexlab{b}}.
\newblock \href {https://doi.org/10.18653/v1/2022.acl-long.498} {Distantly supervised named entity recognition via confidence-based multi-class positive and unlabeled learning}.
\newblock In \emph{Proceedings of the 60th Annual Meeting of the Association for Computational Linguistics (Volume 1: Long Papers)}, pages 7198--7211, Dublin, Ireland. Association for Computational Linguistics.

\bibitem[{Zhou et~al.(2021)Zhou, Liu, Wang, Zhai, Jiang, and Ji}]{DBLP:conf/iccv/ZhouLWZJJ21}
Xiong Zhou, Xianming Liu, Chenyang Wang, Deming Zhai, Junjun Jiang, and Xiangyang Ji. 2021.
\newblock \href {https://doi.org/10.1109/ICCV48922.2021.00014} {Learning with noisy labels via sparse regularization}.
\newblock In \emph{2021 {IEEE/CVF} International Conference on Computer Vision, {ICCV} 2021, Montreal, QC, Canada, October 10-17, 2021}, pages 72--81. {IEEE}.

\end{thebibliography}
\bibliographystyle{acl_natbib}

\end{document}